\documentclass[10pt,twocolumn,letterpaper]{article}

\usepackage[pagenumbers]{cvpr}

\usepackage{multirow}
\usepackage{microtype}

\newcommand{\method}{PlanDiv\xspace}

\DeclareMathOperator*{\argmax}{arg\,max}

\definecolor{cvprblue}{rgb}{0.21,0.49,0.74}
\usepackage[pagebackref,breaklinks,colorlinks,allcolors=cvprblue]{hyperref}

\def\confName{CVPR}
\def\confYear{2027}

\title{PlanDiv: Test-Time Scaling with Latent Plans\\for Vision-Language-Action Models}

\author{Hanbo Zhang\textsuperscript{1} \quad Lianlei Shan\textsuperscript{2,\dag} \quad Liuhaichen Yang\textsuperscript{1} \quad Zezhi Tang\textsuperscript{1,\dag}\\
\textsuperscript{1}University College London \qquad \textsuperscript{2}Tsinghua University\\
{\small \textsuperscript{\dag}Corresponding authors}}

\begin{document}
\maketitle
\begin{abstract}
Test-time scaling improves vision-language-action (VLA) policies by drawing several action candidates and executing the best one.
We find that its returns diminish quickly when the candidates are resampled from one decoder: they differ by noise rather than by plan, and for each of three resampling sources, going from four to eight candidates adds only 0.06--0.07 to the average number of completed tasks on CALVIN.
We introduce \method, an action head that makes the candidates differ by plan: a lightweight planner proposes $K$ latent plan hypotheses, a shared masked-diffusion executor decodes one action chunk per plan from a single backbone pass, and the policy executes the chunk that its own likelihood ranks highest.
The plans are learned from demonstrations alone, without plan labels or rewards.
With the backbone, data and training recipe held fixed, eight plan-conditioned candidates raise five-task chain success on CALVIN ABC$\rightarrow$D from 56.9\% to 69.9\%, while eight candidates from autoregressive, plan-free or same-plan resampling reach at most 64.9\%.
The same-plan baseline is \method's own checkpoint with the plan held fixed, so this gap comes from the plans, not from drawing more samples.
On LIBERO-Long, four plans remove 40\% of single-candidate failures, against at most 21\% for matched resampling.
On a real Franka arm the gain appears where a task admits more than one solution: \method is on par with a single-output OpenVLA-OFT+ baseline on a one-solution task and succeeds in 51 of 54 trials against 40 of 54 on two tasks with two solutions.
\end{abstract}

\section{Introduction}
\label{sec:intro}

Vision-language-action (VLA) models map observations and an instruction to robot actions with a single pretrained backbone~\cite{rt1,rt2,openvla}, and the optimized fine-tuning (OFT) recipe made them fast through parallel decoding and action chunking~\cite{oft}.
Test-time scaling has recently begun to improve these policies without retraining them: instead of executing the policy's single output, the robot draws several action candidates and executes the best one.
RoboMonkey samples $N$ actions and ranks them with a learned verifier~\cite{robomonkey}, RoVer adds a process reward model~\cite{rover}, MG-Select removes the verifier and ranks samples by their divergence from a condition-masked reference~\cite{mgselect}, and CoVer widens the pool by rephrasing the instruction~\cite{cover}.

\begin{figure}[t]
\centering
\includegraphics[width=\linewidth]{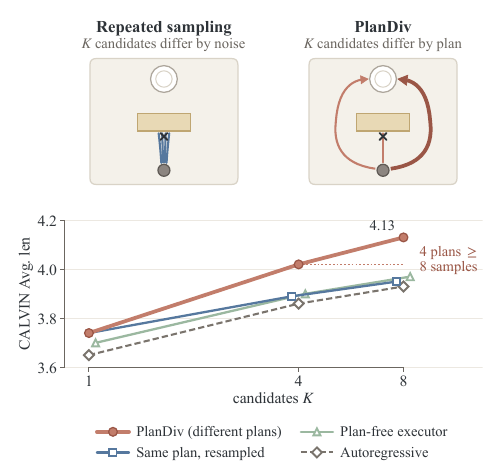}
\caption{Eight plan-conditioned candidates outperform eight resampled ones, and four plans complete more tasks than eight samples from any resampling source. Top (illustration; circle: object, box: obstacle, ring: goal): resampling one decoder yields candidates that differ by noise, so they share one approach and are blocked together, whereas \method decodes one candidate per learned plan, so the candidates take different routes. Bottom: Avg.~len (average number of tasks completed in a five-task chain) on CALVIN ABC$\rightarrow$D, all heads trained under one recipe; same plan, plan-free executor and autoregressive are same-z, executor-only and OpenVLA-AR in Table~\ref{tab:scaling}.}
\label{fig:teaser}
\end{figure}

These methods differ in how candidates are scored, but they share where the candidates come from: one decoder, queried repeatedly through temperature sampling, action perturbation or a rephrased instruction.
We measure what this source buys under one training recipe (Sec.~\ref{sec:diagnosis}), turning what prior work has observed~\cite{mgselect} into a controlled comparison.
Resampling an autoregressive head, a plan-free masked-diffusion head, or a plan-conditioned head held to one fixed plan gives nearly the same curve: most of the gain arrives with the first four candidates, and the next four add only 0.06--0.07 completed tasks on CALVIN (Fig.~\ref{fig:teaser}).
The candidates themselves show why.
Candidates resampled under one plan differ in 18\% of their tokens, and even an oracle selector over eight of them succeeds in only 86\% of sub-task start states, against 76\% for a single candidate.
If the policy commits to a wrong approach direction or sub-goal order, its samples tend to inherit that commitment, and no scorer can select a behavior that the candidate set does not contain.

\begin{figure*}[t]
\centering
\includegraphics[width=\textwidth]{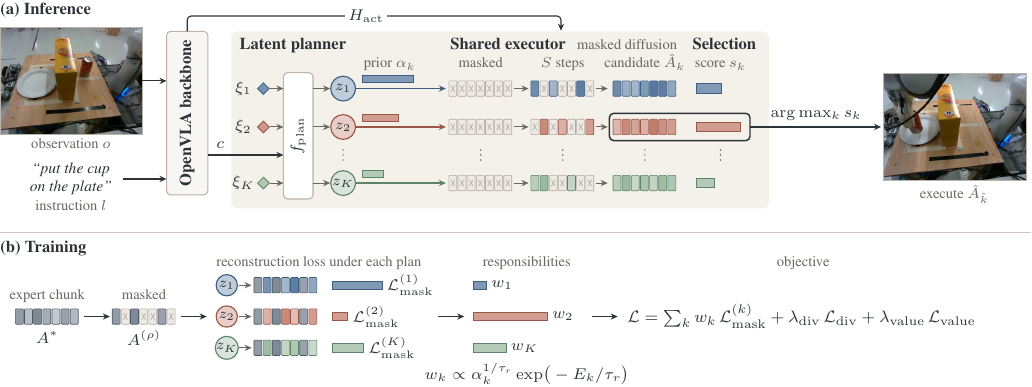}
\caption{Overview of \method. \textbf{(a) Inference.} One backbone pass gives the hidden states at the action positions, $H_{\mathrm{act}}$, and a pooled context $c$. The latent planner turns $K_{\max}$ learned seeds $\xi_k$ into plan hypotheses $z_k$ with a prior $\alpha_k$; only the top-$K$ by prior are decoded, so one checkpoint serves any budget $K\le K_{\max}$. One shared masked-diffusion executor decodes a candidate chunk $\hat A_k$ under each plan in $S{=}4$ unmasking steps, so the candidates come from $K$ different plans instead of $K$ samples under one condition. The policy executes the candidate with the highest score, by default the mean log-probability of its tokens. \textbf{(b) Training.} The masked reconstruction loss of a demonstration chunk under each plan gives soft responsibilities $w_k$ that weight the execution loss; no plan labels or rewards are used. The shaded region is the head we add; the backbone is inherited from OpenVLA. The photographs are frames of our robot setup and are illustrative; token strips and bars are schematic.}
\label{fig:overview}
\end{figure*}

We introduce \method (for \emph{plan diversity}; Fig.~\ref{fig:overview}), an action head that makes the candidates differ by plan instead of by noise: it decodes one candidate per plan and lets the policy choose among them instead of averaging the plans.
A lightweight planner turns $K$ learned seeds into $K$ latent plan hypotheses with a prior over them; a shared masked-diffusion executor, inherited from recent discrete-diffusion action decoders~\cite{ddvla,udvla,mmadavla}, decodes one action chunk per plan from a single backbone pass; and the policy executes the chunk it scores highest.
Planner and executor are trained jointly from demonstrations alone by \emph{latent responsibility assignment}, an objective in the style of soft expectation-maximization (EM) that credits each demonstration chunk to the hypotheses that explain it best, rather than to all of them equally or to a single winner.
The OpenVLA backbone, tokenizer and action-token interface are unchanged, and one checkpoint serves any candidate budget $K\le K_{\max}$.

With the backbone, data, training recipe and selector matched, eight plan-conditioned candidates raise CALVIN five-task chain success from 56.9\% to 69.9\%, against 63.9--64.9\% for the resampling sources.
Same-plan resampling shares the checkpoint, the first candidate and each candidate's random numbers with \method, so the gap between them isolates the effect of the plans.
Eight plan-conditioned candidates remove 30\% of single-candidate failures on CALVIN and four remove 40\% on LIBERO-Long, against 17--21\% for matched resampling and 5--18\% that prior test-time scaling reports on other base policies~\cite{rover,robomonkey,mgselect}.
Four plans are already worth eight samples: they complete more tasks than eight resampled candidates from any of the three sources, and they cover more start states (88.8\% against 86.1\% for eight same-plan samples).
Changing only the plan under fixed decoder noise alters the task outcome more than twice as often as changing only the noise. In simulation, eight candidates take 288\,ms, about a third of the time an autoregressive policy needs for one.

Our contributions are:
\begin{itemize}
  \item \textbf{A controlled measurement.} With the backbone, data, training recipe and selector matched, resampling one decoder has quickly diminishing returns: three resampling sources gain only 0.06--0.07 completed tasks from four to eight candidates, and even an oracle selector over eight same-plan resamples succeeds in only 86\% of sub-task start states.
  \item \textbf{\method.} An action head that decodes plan-conditioned candidates, one per latent plan hypothesis, from one backbone pass; the plans are learned from demonstrations alone by latent responsibility assignment, with no plan labels or rewards.
  \item \textbf{Evidence.} At matched candidate count \method outperforms all three resampling sources on CALVIN and LIBERO-Long, including same-plan resampling from the same checkpoint; a fixed-noise intervention shows that the plan alone changes task outcomes, and without soft responsibilities or repulsion between plans the gain falls to that of plain resampling. On a real Franka arm, on tasks with more than one solution, \method removes 11 of the 14 failures of an OpenVLA-OFT+ baseline.
\end{itemize}

\section{Related Work}
\label{sec:related}

\noindent\textbf{Vision-language-action models.}
VLAs fine-tune a pretrained vision-language model to output robot actions~\cite{rt2,openvla}.
Recent models mostly change what the backbone produces before or alongside the action: spatial representations~\cite{spatialvla}, visual traces~\cite{tracevla}, visual chains of thought~\cite{cotvla}, future observations~\cite{seer,worldvla,unifiedvla}, or task-centric latent actions learned from video~\cite{univla}.
Others make the policy smaller~\cite{nora,vlaadapter}, replace token decoding with flow matching or diffusion~\cite{pizero,flower,diffusionvla}, compress the action sequence~\cite{fast}, or keep improving a trained policy through interactive post-training~\cite{riptvla}.
Each of them produces one action or action chunk per decision.
\method keeps the backbone and changes only the action head and the candidates it produces at test time.

\noindent\textbf{VLA action decoders.}
We build on the OpenVLA family: a pretrained vision-language backbone with a discrete action-token interface~\cite{openvla,rt2}, adapted with parallel decoding and action chunking~\cite{oft,pdvla}.
Our executor belongs to the recent line of masked-diffusion action decoders, which decode a chunk by iterative unmasking rather than left-to-right generation~\cite{ddvla,udvla,mmadavla}, following D3PM, MaskGIT and MDLM~\cite{d3pm,maskgit,mdlm}.
We keep the decoder standard so that the comparison isolates the plan it is conditioned on.
Non-autoregressive translation met the same multimodality problem in parallel decoding: it conditions the decoder on latent fertilities obtained from an external aligner and, at test time, decodes several of them and rescores the outputs with an autoregressive teacher~\cite{nat}.
Our plans are learned without supervision and ranked by the executor's own likelihood.

\noindent\textbf{Test-time scaling for VLAs.}
Most prior work improves how candidates are selected.
RoboMonkey draws $N$ samples by temperature or Gaussian perturbation and ranks them with a 7B verifier~\cite{robomonkey}; RoVer adds a process reward model with direction-guided expansion~\cite{rover}; MG-Select removes the verifier and ranks by the KL divergence to a condition-masked reference~\cite{mgselect}; ELASTIC learns how to split compute between denoising steps and parallel samples~\cite{elastic}.
V-GPS re-ranks the actions of a generalist policy with a value function learned by offline reinforcement learning~\cite{vgps}, and Bidirectional Decoding selects among sampled action chunks by their coherence with past decisions~\cite{bid}.
CoVer changes the input instead, rephrasing the instruction and scoring instruction--action pairs with a contrastive verifier~\cite{cover}.
\method changes the source of the candidates: its hypotheses are learned from demonstrations alone, and each one is decoded into its own candidate, so that the policy chooses among alternatives rather than averaging them.
It is orthogonal in design to the selection-side methods, whose scores operate on whatever candidates they are given.

\noindent\textbf{Multiple hypotheses and latent plans.}
Expressive policy classes capture multimodal demonstrations within one conditional distribution, through discretized action modes~\cite{bet,vqbet}, a conditional variational autoencoder (CVAE) latent~\cite{act} or diffusion~\cite{diffusionpolicy}; drawing several samples from it is the resampling we analyze in Sec.~\ref{sec:diagnosis}.
Play-LMP and HULC learn a latent plan space from play data and condition a local policy on one sampled plan~\cite{playlmp,hulc}; hierarchical and skill-based methods use temporally extended abstractions in the same spirit~\cite{hiro,hac,spirl,latentskillspace}.
Choice Policies predicts $K$ candidate action sequences in one forward pass together with a learned score, trained winner-takes-all in the multiple-choice-learning tradition to model multimodal behavior with fast inference~\cite{choicepolicy,mcl2012,smcl}; relaxed and evolving variants soften the winner-takes-all assignment~\cite{rwta,ewta}.
Latent responsibility assignment is a soft, prior-weighted counterpart of that objective, an EM-style step induced by the masked reconstruction loss~\cite{em}, and our $K$ hypotheses share one executor rather than owning $K$ heads; sharpening our assignment toward winner-takes-all reduces the gain from extra candidates to that of resampling (Sec.~\ref{sec:structure}).
MoDE routes among expert denoisers by noise level~\cite{mode}, whereas our hypotheses are alternatives for the same decision; unlike CVAE policies~\cite{cvae,act}, we amortize no posterior and the planner proposes candidates explicitly.
DLow maps one random variable through $K$ learned transforms into diverse latent codes of a pretrained generator~\cite{dlow}; our hypotheses are trained jointly with the executor and carry a prior that ranks them at test time.
In language modeling, Mode-Conditioning allocates samples evenly across modes obtained from labels or gradient clustering~\cite{modc}, PlanSearch searches over natural-language plans because sampled programs repeat one another's mistakes~\cite{plansearch}, Large Language Monkeys separates coverage from selection~\cite{monkeys}, and SoftCoT++ diversifies a latent thought through several learned initial tokens~\cite{softcotpp}.
These settings count a problem as solved if any sample is correct; a robot must commit to one candidate before its outcome is known, so \method computes its plans from the current context, learns them jointly with the executor, and ranks them with a learned prior.

\section{The Diminishing Returns of Resampling}
\label{sec:diagnosis}

Before changing the policy, we measure what repeated sampling delivers.
Three resampling sources come from heads trained from the same initialization with one recipe (Sec.~\ref{sec:setup}): the discrete autoregressive head of OpenVLA, sampled with temperature (\emph{OpenVLA-AR}); a plan-free masked-diffusion head, sampled $K$ times (\emph{executor-only}); and the plan-conditioned head of Sec.~\ref{sec:method} with every candidate decoded under its top-prior plan (the \emph{same-z} source, $z$ being the plan).
All three rank their candidates by the same likelihood score and execute the best one.

\noindent\textbf{Returns diminish after four candidates.}
On CALVIN ABC$\rightarrow$D the three sources gain 0.15--0.21 completed tasks from one to four candidates and only 0.06--0.07 from four to eight (Table~\ref{tab:scaling}, Fig.~\ref{fig:teaser}).
The three heads differ in decoder or plan conditioning, yet their gains almost coincide, which points to what they share: every candidate is a draw from one conditional distribution.
A published sweep has the same shape: MG-Select reports only marginal improvement beyond four samples, with 48.1\% success on RoboCasa at $N{=}4$ and 48.4\% at $N{=}64$~\cite{mgselect}.

\noindent\textbf{The candidate pool bounds even a perfect selector.}
On a bank of CALVIN sub-task start states, where every candidate is executed and receives a real outcome (Sec.~\ref{sec:mechanism}), candidates resampled under one plan differ in 18\% of their 56 token positions.
A single candidate succeeds in 76\% of the states; with eight resamples at least one succeeds in 86\% (Table~\ref{tab:mechanism}a).
An oracle that always picked a successful candidate would therefore recover 41.6\% of the first-candidate failures and no more: the pool, not the selector, sets the ceiling, so we change where the candidates come from.

\noindent\textbf{Why the curve flattens.}
Suppose a start state admits several approaches, and a candidate decoded under approach $m$ succeeds with probability $q_m$.
If all $K$ candidates follow the approach the policy has committed to, $m^\star$, at least one succeeds with probability $C_K=1-(1-q_{m^\star})^K$.
Each extra candidate retries the same approach: doubling the budget adds $C_{2K}-C_K=(1-q_{m^\star})^K C_K$, which shrinks geometrically, and a state that $m^\star$ cannot solve ($q_{m^\star}{=}0$) is never recovered.
If instead the $K$ candidates follow $K$ different approaches, the probability is $1-\prod_{k=1}^{K}(1{-}q_{m_k})$, and such a state is recovered as soon as one included approach can solve it.
Resampling behaves like the first case; \method is built to produce the second, in which the curve still flattens but toward a higher ceiling.

\section{Method}
\label{sec:method}

To change where the candidates come from (Sec.~\ref{sec:diagnosis}), \method replaces the action head (Fig.~\ref{fig:overview}).
Sec.~\ref{sec:base} states the interface we inherit, Sec.~\ref{sec:mode} the single change we make to it, Secs.~\ref{sec:planner}--\ref{sec:executor} the two modules that realize the change, Sec.~\ref{sec:resp} how they are trained without plan labels, and Sec.~\ref{sec:inference} what happens at test time.

\subsection{OpenVLA action-token interface}
\label{sec:base}
Given observations $o$ and instruction $l$, the OpenVLA backbone produces hidden states at the $T$ action-token positions, $H_{\mathrm{act}}=\{h_i\}_{i=1}^{T}=f_{\mathrm{ctx}}(o,l)$, with $T=DH$ for an $H$-step chunk of $D$-dimensional actions and pooled context $c=\frac{1}{T}\sum_i h_i$.
Continuous actions are normalized with dataset statistics and digitized into the $B{=}256$ uniform bins of the OpenVLA tokenizer, so a chunk is $A\in\{0,\ldots,B-1\}^{T}$.
The autoregressive interface we inherit models the chunk as
\begin{equation}
  p_{\mathrm{AR}}(A \mid H_{\mathrm{act}})=\prod_{i=1}^{T}p_\theta(A_i \mid H_{\mathrm{act}}, A_{<i}),
  \label{eq:ar}
\end{equation}
and the OFT recipe decodes the $T$ positions in parallel with an L1 head~\cite{oft}.
The candidates that most prior test-time scaling methods compare are samples or perturbations of one such decoder.

\subsection{Plan-conditioned candidates}
\label{sec:mode}
We replace the single decoder by a mixture of $K$ plan-conditioned decoders that share all parameters,
\begin{equation}
  p(A \mid o,l)\approx \sum_{k=1}^K \alpha_k(c)\,p_\phi(A \mid H_{\mathrm{act}},z_k),
  \label{eq:mixture}
\end{equation}
where $z_k$ is the $k$th latent plan hypothesis, $\alpha_k(c)$ its learned prior, and $p_\phi$ one shared masked-diffusion executor.
At test time we decode one candidate under each $z_k$ and compare them, instead of drawing $K$ samples under one condition.
All $K_{\max}$ hypotheses are trained; at test time only the top-$K$ by prior are decoded, for any $K\le K_{\max}$ (Sec.~\ref{sec:inference}).

\subsection{Latent plan hypotheses}
\label{sec:planner}
To obtain $K$ distinct conditions from one context, the planner maps $K$ learned seeds $\xi_k$ through a shared network,
\begin{equation}
\begin{aligned}
  z_k &= f_{\mathrm{plan}}(c,\xi_k), \\
  \alpha_k &=\operatorname{softmax}_k\big(g_{\mathrm{prior}}(z_1),\ldots,g_{\mathrm{prior}}(z_K)\big),
\end{aligned}
  \label{eq:planner}
\end{equation}
where $f_{\mathrm{plan}}$ and $g_{\mathrm{prior}}$ are small MLPs and the prior lets the model keep unlikely alternatives without weighting them equally.
A repulsion term on the normalized latents $\bar z_k$ discourages duplicate hypotheses for $K\ge 2$:
\begin{equation}
  \mathcal{L}_{\mathrm{div}}=\frac{1}{K(K-1)}\sum_{k\ne j}(\bar z_k^\top \bar z_j)^2.
  \label{eq:div}
\end{equation}
We use latent rather than symbolic plans because demonstration corpora carry no plan annotations, and because the plans must live in the space the executor can act on.

\subsection{Shared masked-diffusion executor}
\label{sec:executor}
The executor operates directly on action tokens.
During training a mask ratio $\rho\sim\mathcal{U}[0.15,1.0]$ selects a set $M_\rho$ of positions of the expert chunk $A^*$ to replace with a learned mask token, giving the corrupted chunk $A^{(\rho)}$, and the executor is trained to reconstruct them.
With $\ell_i^{(k)}=-\log p_\phi(A_i^* \mid A^{(\rho)}, H_{\mathrm{act}}, z_k)$ the loss at position $i$ under plan $k$,
\begin{equation}
  \mathcal{L}_{\mathrm{mask}}^{(k)} =
  \mathbb{E}_{\rho, M_\rho}\Big[\frac{1}{|M_\rho|}\sum_{i\in M_\rho}\ell_i^{(k)}\Big].
  \label{eq:mask}
\end{equation}
At each position the executor sees the backbone state $h_i$, the current token embedding, the plan $z_k$, a mask-ratio embedding and a position embedding, and predicts a categorical distribution over the $B$ bins.
The positions are conditionally independent given $(H_{\mathrm{act}},z_k)$, so the plan is the only chunk-level coupling the head introduces.
This is deliberately the simplest member of the masked-diffusion decoder family~\cite{ddvla}; it keeps the OpenVLA tokenizer, vocabulary and hidden-state interface intact.

\subsection{Latent responsibility assignment}
\label{sec:resp}
No label says which hypothesis should explain a demonstration.
To let the model decide, we weight the per-plan losses by soft responsibilities computed from prior plausibility and reconstruction quality. With $E_k=\sum_{i\in M_\rho}\ell_i^{(k)}$ the negative log-likelihood of the masked part of a chunk under plan $k$ and $\operatorname{sg}[\cdot]$ the stop-gradient,
\begin{equation}
\begin{aligned}
  w_k &=
  \frac{\exp\!\big((\log \alpha_k-\operatorname{sg}[E_k])/\tau_r\big)}
  {\sum_{j=1}^K \exp\!\big((\log \alpha_j-\operatorname{sg}[E_j])/\tau_r\big)}, \\
  \mathcal{L}_{\mathrm{exec}} &= \sum_{k=1}^K w_k \,\mathcal{L}_{\mathrm{mask}}^{(k)},
\end{aligned}
  \label{eq:responsibility}
\end{equation}
with $\tau_r{=}0.5$ unless stated otherwise.
These are tempered posteriors over plans, $w_k\propto\big(\alpha_k\,p_\phi(A^*_{M_\rho}\mid z_k)\big)^{1/\tau_r}$: they sum the log-likelihood over the masked positions rather than averaging it, so differences between plans are not divided by $|M_\rho|$. They play the role of the E-step in an EM-style procedure~\cite{em}: the executor improves most under the hypotheses that already explain a chunk well, and the prior learns to favor those hypotheses.
Compared with winner-takes-all multiple-choice learning as used by Choice Policies~\cite{choicepolicy,mcl2012}, the assignment is soft and prior-weighted, so a chunk that two hypotheses explain equally keeps training both; compared with a uniform assignment, it lets hypotheses specialize.
Sec.~\ref{sec:structure} tests both alternatives.
The full objective is $\mathcal{L}=\mathcal{L}_{\mathrm{exec}}+\lambda_{\mathrm{div}}\mathcal{L}_{\mathrm{div}}+\lambda_{\mathrm{value}}\mathcal{L}_{\mathrm{value}}$ with $\lambda_{\mathrm{div}}{=}0.01$ and $\lambda_{\mathrm{value}}{=}0.1$, where $\mathcal{L}_{\mathrm{value}}$ trains the value proxy below.

\subsection{Inference and candidate selection}
\label{sec:inference}
At test time the policy runs the backbone once, forms $K$ plans, and for each plan starts from a fully masked chunk and iteratively fills the most confident positions while re-masking the rest for $S{=}4$ steps, producing $K$ candidate chunks $\hat A_1,\ldots,\hat A_K$.
It then commits to the candidate with the highest score.

\noindent\textbf{Selection.}
The default \emph{likelihood score} is the mean log-probability of the sampled tokens of a candidate, recorded as each position is filled.
It needs no additional model, so it applies unchanged to any candidate source, and we use it in all controlled comparisons unless another selector is named.
\method also trains a value proxy $V_\psi$ on expert chunks to predict the negative reconstruction loss under each plan,
\begin{equation}
  \mathcal{L}_{\mathrm{value}}=\frac{1}{K}\sum_k\big(V_\psi(c,z_k,A^*)+\mathcal{L}^{(k)}_{\mathrm{mask}}\big)^2,
  \label{eq:value}
\end{equation}
which gives the full score
\begin{equation}
  s_k = V_\psi(c,z_k,\hat A_k)+\bar\gamma_k+\log\alpha_k,
  \label{eq:selector}
\end{equation}
where $\bar\gamma_k$ is the mean modal token probability of candidate $k$; with either score as $s_k$, the policy executes $\hat A_{\hat k}$ with $\hat k=\argmax_k s_k$.
Sec.~\ref{sec:selector} compares the two scores and their individual terms.

\noindent\textbf{One checkpoint, any $K$.}
Candidate count is a test-time budget, and the prior decides how it is spent.
A model trained with $K_{\max}$ seeds is evaluated at any $K\le K_{\max}$ by decoding only the $K$ hypotheses with the highest prior $\alpha_k(c)$ at the current decision, so the candidate sets for $K{=}1,2,4,8$ are nested prefixes of one ranking and no retraining is involved.

\section{Experiments}
\label{sec:experiments}

\begin{table*}[t]
\centering
\footnotesize
\setlength{\tabcolsep}{5pt}
\begin{tabular}{@{}lcccccccc@{}}
\toprule
 & & \multicolumn{6}{c}{CALVIN ABC$\rightarrow$D} & LIBERO \\
\cmidrule(lr){3-8}\cmidrule(l){9-9}
 & & \multicolumn{5}{c}{Tasks completed in a row (\%)} & & \\
\cmidrule(lr){3-7}
Candidate source & $K$ & 1 & 2 & 3 & 4 & 5 & Avg.~len & Long (\%) \\
\midrule
OpenVLA-AR & 1 & 92.8 & 82.9 & 72.2 & 62.3 & 54.8 & 3.65 & 84.8 \\
 & 4 & 94.3 & 85.8 & 76.6 & 67.4 & 61.9 & 3.86 & 87.6 \\
 & 8 & 94.9 & 87.2 & 77.8 & 69.2 & 63.9 & 3.93 & -- \\
\midrule
Executor-only & 1 & 93.4 & 83.8 & 73.6 & 63.1 & 56.1 & 3.70 & 85.8 \\
 & 4 & 94.7 & 86.4 & 77.6 & 68.4 & 62.9 & 3.90 & 88.8 \\
 & 8 & 95.2 & 87.4 & 79.2 & 70.3 & 64.9 & 3.97 & -- \\
\midrule
\method (same-z) & 4 & 94.8 & 86.3 & 77.1 & 68.7 & 62.1 & 3.89 & 89.2 \\
 & 8 & 95.1 & 87.2 & 78.6 & 69.7 & 64.4 & 3.95 & -- \\
\midrule
\method (different-z) & 1 & 93.9 & 84.8 & 74.3 & 64.1 & 56.9 & 3.74 & 86.4 \\
 & 4 & 95.6 & 88.8 & 80.2 & 72.3 & 65.1 & 4.02 & \textbf{91.8} \\
 & 8 & \textbf{96.2} & \textbf{89.8} & \textbf{82.2} & \textbf{74.9} & \textbf{69.9} & \textbf{4.13} & -- \\
\bottomrule
\end{tabular}
\caption{Eight plan-conditioned candidates outperform every resampling source at the same budget, and four score higher than eight resampled ones in every CALVIN column. Candidate source $\times$ candidate count under the likelihood selector, all heads trained under one recipe. CALVIN ABC$\rightarrow$D: tasks completed in a row (\%) and Avg.~len on 1000 chains. LIBERO-Long: success on 500 episodes; all sources are evaluated at $K\in\{1,4\}$, the range of the $K_{\max}{=}4$ \method checkpoint, and -- marks settings not evaluated. Same-z equals different-z at $K{=}1$ (shared first candidate) and is listed once.}
\label{tab:scaling}
\end{table*}

Four questions organize this section: whether plan-conditioned candidates beat resampled ones at equal budget (Sec.~\ref{sec:scaling}), whether the plan is the cause (Sec.~\ref{sec:mechanism}), what selection and the training objective contribute and what the candidates cost (Secs.~\ref{sec:selector} and~\ref{sec:structure}), and whether on a real robot the gain appears where a task admits more than one solution (Sec.~\ref{sec:real}).

\subsection{Setup}
\label{sec:setup}
\noindent\textbf{Benchmarks.}
CALVIN ABC$\rightarrow$D~\cite{calvin} trains on environments A, B and C and evaluates in the unseen environment D on 1000 five-task instruction chains; we report the success rate of completing at least 1--5 tasks in a row and the average number of completed tasks (Avg.~len).
LIBERO-Long~\cite{libero} has 10 long-horizon tasks, each evaluated on 50 initial states (500 episodes of at most 520 steps).
All models use the OpenVLA-7B backbone with LoRA~\cite{lora} of rank 32 and 8-step chunks of 7-dimensional actions ($T{=}56$ tokens, $B{=}256$ bins).

\noindent\textbf{Matched comparison.}
Four candidate sources, obtained from three heads trained from the same initialization with one recipe (AdamW, batch size 64, 40{,}000 steps, two $224{\times}224$ views and proprioception), are compared at equal candidate count and equal executed chunk length:
\emph{(i)} \method, which decodes each candidate under a different plan (\emph{different-z});
\emph{(ii)} \emph{same-z}, the same \method checkpoint with every candidate decoded under the prior's top plan, so that candidates differ only by decoder noise; it shares the first candidate and each candidate's random numbers with \emph{(i)};
\emph{(iii)} \emph{executor-only}, the same executor trained without the planner or plan conditioning, sampled $K$ times;
\emph{(iv)} \emph{OpenVLA-AR}, the discrete autoregressive head of Eq.~\eqref{eq:ar} trained with the same recipe, whose first candidate is greedy and whose other candidates are temperature-sampled, with the temperature chosen on a validation set.
Unless another selector is named, every source is ranked by the likelihood score of Sec.~\ref{sec:inference}.
\method decodes with $S{=}4$ unmasking steps at temperature $1.0$ and is trained with $K_{\max}{=}8$ hypotheses on CALVIN and $K_{\max}{=}4$ on LIBERO-Long.

\subsection{Plans versus samples}
\label{sec:scaling}

\noindent\textbf{At equal budget.}
At $K{=}8$ \method reaches 4.13 (Table~\ref{tab:scaling}, Fig.~\ref{fig:teaser}), against 3.97 for executor-only, 3.95 for same-z and 3.93 for OpenVLA-AR, a margin of $+0.16$ to $+0.20$, and it is higher in every chain column.
At $K{=}4$ on LIBERO-Long it reaches 91.8\%, against 88.8\% (executor-only), 89.2\% (same-z) and 87.6\% (OpenVLA-AR).
Because same-z and different-z share the checkpoint, the first candidate and each candidate's random numbers, their difference isolates the plan variable: resampling eight times under one plan adds $+0.21$ over one candidate ($3.74\rightarrow3.95$), whereas decoding under eight different plans adds $+0.39$ ($3.74\rightarrow4.13$), nearly double.
Same-z is a strong control: executor-only, trained without plans, ends at the same level (3.97).
Eight plan-conditioned candidates remove 30\% of the single-candidate policy's five-task-chain failures ($43.1\%\rightarrow30.1\%$), against 17--20\% for the three resampling sources; on LIBERO-Long four candidates remove 40\% ($13.6\%\rightarrow8.2\%$), against 18--21\%.

\begin{table}[t]
\centering
\footnotesize
\setlength{\tabcolsep}{3pt}
\resizebox{\columnwidth}{!}{%
\begin{tabular}{@{}llccc@{}}
\toprule
Method & Extra model & Base policy & Before$\,\rightarrow\,$after & Removed (\%) \\
\midrule
\multicolumn{5}{@{}l}{\emph{CALVIN ABC$\rightarrow$D, five-task chains (\%)}} \\
RoVer~\cite{rover} & reward model & GR-1~\cite{gr1} & 43.4$\,\rightarrow\,$48.7 & 9 \\
RoVer~\cite{rover} & reward model & Dita~\cite{dita} & 50.0$\,\rightarrow\,$59.2 & 18 \\
RoVer~\cite{rover} & reward model & MoDE~\cite{mode} & 63.5$\,\rightarrow\,$66.6 & 8 \\
\method ($K{=}8$) & none & \method ($K{=}1$) & 56.9$\,\rightarrow\,$69.9 & \textbf{30} \\
\midrule
\multicolumn{5}{@{}l}{\emph{LIBERO-Long (\%)}} \\
RoboMonkey~\cite{robomonkey} & 7B verifier & OpenVLA~\cite{openvla} & 49.8$\,\rightarrow\,$56.5 & 13 \\
MG-Select~\cite{mgselect} & none & OpenVLA~\cite{openvla} & 52.2$\,\rightarrow\,$54.7 & 5 \\
MG-Select~\cite{mgselect} & none & $\pi_0$-FAST~\cite{fast} & 79.6$\,\rightarrow\,$82.7 & 15 \\
\method ($K{=}4$) & none & \method ($K{=}1$) & 86.4$\,\rightarrow\,$91.8 & \textbf{40} \\
\bottomrule
\end{tabular}}
\caption{\method removes a larger share of the base policy's failures than RoVer, RoboMonkey and MG-Select report on the same benchmarks and, like MG-Select, needs no extra model. ``Removed'' is the share of the base policy's failures that test-time scaling eliminates. Base policies differ across methods, so the table compares the gain from scaling, not absolute performance.}
\label{tab:tts}
\end{table}

Table~\ref{tab:tts} sets these gains beside published test-time scaling results on the same benchmarks and metrics.
These methods, which resample a policy and rescore the samples with or without an external model, remove 5--18\% of the base policy's failures.
\method removes 30\% and 40\% without any external model.

\noindent\textbf{Four plans are worth eight samples.}
With four candidates \method reaches an Avg.~len of 4.02, above what any resampling source reaches with eight (3.93--3.97), is higher in every chain column, and takes less time than eight executor-only samples (188 against 257\,ms, Table~\ref{tab:cost}).
From four to eight candidates it gains another 0.11, where the three resampling sources gain 0.06--0.07; overall eight candidates add $+0.39$ to \method and $+0.21$ to $+0.28$ to resampling.

\subsection{Isolating the effect of the plan}
\label{sec:mechanism}

\begin{table}[t]
\centering
\footnotesize
\setlength{\tabcolsep}{4pt}
\begin{tabular}{@{}lcc@{}}
\toprule
(a) Candidate pools & same-z & different-z \\
\midrule
Token Hamming distance & 0.182 & 0.304 \\
End-effector spread & 0.064 & 0.107 \\
Branch success, likelihood (\%) & 77.8 & 83.2 \\
Cover@1 / @2 (\%) & 76.2 / 80.2 & 76.2 / 83.4 \\
Cover@4 / @8 (\%) & 83.2 / 86.1 & 88.8 / 92.1 \\
First-candidate failures recovered (\%) & 41.6 & 66.8 \\
\bottomrule
\end{tabular}

\vspace{4pt}
\begin{tabular}{@{}lccc@{}}
\toprule
(b) Intervention & RMSE & Changed (\%) & Outcome (\%) \\
\midrule
same plan, same noise & 0.000 & 0.0 & 0.0 \\
same plan, new noise & 0.041 & 71.8 & 7.2 \\
new plan, same noise & 0.092 & 91.2 & 16.2 \\
new plan, new noise & 0.108 & 94.8 & 18.8 \\
\bottomrule
\end{tabular}
\caption{Candidates decoded under different plans are farther apart and cover more states, and changing only the plan changes outcomes more than twice as often as changing only the noise. (a) Same-z versus different-z candidates at $K{=}4$ (Cover@$K$ up to $K{=}8$). Cover@$K$ is the share of states where at least one of the $K$ candidates leads to sub-task success; the last row is $(\text{Cover@8}-\text{Cover@1})/(1-\text{Cover@1})$. (b) Fixed-noise intervention: RMSE on normalized actions; ``changed'' means at least one token differs; ``outcome'' is the share of states whose sub-task outcome differs.}
\label{tab:mechanism}
\end{table}

The closed-loop results show that plan-conditioned candidates are more useful; here we test whether they are more different, and whether the plan alone changes what is executed.
We build a bank of sub-task start states from 200 reference chains fixed in advance.
From each state, every candidate chunk is executed for 8 steps after resetting to the state, and a frozen reference policy then continues until the sub-task succeeds or 360 steps have elapsed, so each candidate (a branch) receives a real outcome.

\noindent\textbf{Coverage.}
Different-z candidates are farther apart than same-z candidates (Table~\ref{tab:mechanism}a) and cover more states.
Both sources share the first candidate, which succeeds in 76.2\% of the states.
With eight candidates, at least one succeeds in 92.1\% of the states under different plans and in 86.1\% under one plan: different plans recover 66.8\% of the first-candidate failures and resampling recovers 41.6\%.
Four plans already cover more states (88.8\%) than eight same-plan samples.
At $K{=}4$, the likelihood selector converts part of the coverage into branch success, the share of states where the selected candidate succeeds: 83.2\% against 77.8\%.

\noindent\textbf{Fixed-noise intervention.}
To attribute the difference to the plan, we fix the observation, the model state, every unmasking random number, the mask order and the simulator state, and replace only the plan, only the decoder noise, or both (Table~\ref{tab:mechanism}b).
Replacing only the plan changes the task outcome in 16.2\% of the states, more than twice as often as replacing only the noise (7.2\%), and moves the actions more than twice as far (RMSE 0.092 against 0.041).
The plan alone, with all randomness fixed, can thus change what is executed and how the sub-task ends.

\subsection{Selection and cost}
\label{sec:selector}

\begin{table}[t]
\centering
\footnotesize
\setlength{\tabcolsep}{4pt}
\begin{tabular}{@{}lcc@{}}
\toprule
Selector ($K{=}4$) & Avg.~len & Branch success (\%) \\
\midrule
first (top prior; equals $K{=}1$) & 3.74 & 76.2 \\
random & 3.71 & 74.8 \\
confidence $\bar\gamma_k$ & 3.96 & 80.8 \\
likelihood (default) & 4.02 & 83.2 \\
value $V_\psi$ & 3.99 & 82.2 \\
full score, Eq.~\eqref{eq:selector} & \textbf{4.08} & \textbf{84.8} \\
\midrule
oracle (any branch succeeds) & -- & 88.8 \\
\bottomrule
\end{tabular}
\caption{The likelihood score, which needs no extra model, captures most of the gain of selection, and the full score adds 0.06. Selectors for \method at $K{=}4$: closed-loop Avg.~len on 1000 CALVIN chains and branch success on the state bank of Sec.~\ref{sec:mechanism}, where the oracle equals Cover@4.}
\label{tab:selector}
\end{table}

\noindent\textbf{Selection.}
Table~\ref{tab:selector} varies only the selector.
Picking a candidate at random is no better than always taking the first, so the gain of Table~\ref{tab:scaling} does not come from executing a different candidate by chance.
The default likelihood score, which needs no additional model, reaches 4.02; the full score of Eq.~\eqref{eq:selector} raises this to 4.08.
On the state bank the full score reaches 84.8\% branch success against 88.8\% for an oracle: selection already captures most of what the candidates offer.

\begin{table}[t]
\centering
\footnotesize
\setlength{\tabcolsep}{4.5pt}
\begin{tabular}{@{}lccccc@{}}
\toprule
Source & $K$ & p50 (ms) & p95 (ms) & Mem.\ (GB) & Avg.~len \\
\midrule
\method & 1 & 118 & 157 & 18.8 & 3.74 \\
 & 4 & 188 & 248 & 19.3 & 4.02 \\
 & 8 & 288 & 374 & 19.9 & 4.13 \\
\midrule
Executor-only & 1 & 109 & 146 & 18.4 & 3.70 \\
 & 4 & 169 & 223 & 18.8 & 3.90 \\
 & 8 & 257 & 336 & 19.4 & 3.97 \\
\midrule
OpenVLA-AR & 1 & 848 & 1094 & 18.4 & 3.65 \\
 & 4 & 1262 & 1642 & 21.2 & 3.86 \\
 & 8 & 1798 & 2342 & 24.6 & 3.93 \\
\bottomrule
\end{tabular}
\caption{Eight plan-conditioned candidates take 288\,ms, about a third of the time OpenVLA-AR needs for one. End-to-end latency per policy call in simulation (CALVIN; p50: median; p95: 95th percentile) and peak memory on one 80\,GB GPU at batch size 1, with the Avg.~len of Table~\ref{tab:scaling}.}
\label{tab:cost}
\end{table}

\noindent\textbf{Cost.}
Eight plan-conditioned candidates take 288\,ms at the median (Table~\ref{tab:cost}), 31\,ms more than eight executor-only samples and about a third of the 848\,ms that OpenVLA-AR needs for a single chunk (1798\,ms for eight), yet \method completes 0.48 more tasks (4.13 against 3.65).
For a 30\,Hz controller that consumes 8-step chunks, a new chunk is due every 267\,ms: \method meets this at $K{=}4$ (p95 248\,ms), and no OpenVLA-AR configuration does.

\subsection{Credit assignment and the value of extra plans}
\label{sec:structure}

\begin{table}[t]
\centering
\footnotesize
\setlength{\tabcolsep}{3.8pt}
\begin{tabular}{@{}lcccc@{}}
\toprule
Training objective & $K{=}1$ & $K{=}4$ & Gain & Eff.\ plans \\
\midrule
\method & 3.74 & \textbf{4.02} & \textbf{+0.28} & 3.04 \\
No repulsion ($\lambda_{\mathrm{div}}{=}0$) & 3.73 & 3.93 & +0.20 & 1.98 \\
Uniform responsibility ($w_k{=}1/K$) & 3.70 & 3.88 & +0.18 & 3.42 \\
Sharp responsibility ($\tau_r{=}0.05$) & 3.72 & 3.90 & +0.18 & 1.58 \\
\bottomrule
\end{tabular}
\caption{Removing the repulsion term or the soft responsibilities reduces the gain from four candidates to the level of plain resampling. Training-objective ablations on CALVIN ABC$\rightarrow$D (Avg.~len, likelihood selector). ``Gain'' is $K{=}4$ minus $K{=}1$; ``Eff.\ plans'' is the exponentiated entropy of the selection frequency over the four slots (maximum 4).}
\label{tab:ablation}
\end{table}

Table~\ref{tab:ablation} asks what makes extra plans pay off by removing the two ingredients of the objective, soft responsibilities and repulsion.
With a single candidate all variants stay within 0.04 of the full model (3.70--3.73 against 3.74 Avg.~len), so the objective acts on the additional candidates.
Without the repulsion term, with uniform responsibilities, or with responsibilities sharpened toward winner-takes-all ($\tau_r{=}0.05$), the gain from four candidates falls from $+0.28$ to $+0.18$--$0.20$, the level that plain resampling reaches in Table~\ref{tab:scaling} ($+0.15$ to $+0.21$).
Uniform responsibilities spread selection most evenly (3.42 effective plans) yet gain no more than the sharpest assignment (1.58), so evenly used plans are not by themselves useful; the gain needs both the soft, prior-weighted assignment and repulsion.

\subsection{Real-robot evaluation}
\label{sec:real}

\begin{table}[t]
\centering
\footnotesize
\setlength{\tabcolsep}{6pt}
\begin{tabular}{@{}lcc@{}}
\toprule
 & OFT+ & \method \\
\midrule
Push a bowl (one solution) & 24/25 & 25/25 \\
Two blocks into a tray & 17/21 & \textbf{21/21} \\
Can around an obstacle onto a plate & 23/33 & \textbf{30/33} \\
\midrule
Two-solution tasks (\%) & 74.1 & \textbf{94.4} \\
Time per query (ms) & 145 & \textbf{113} \\
\bottomrule
\end{tabular}
\caption{On a real Franka arm \method removes 11 of the baseline's 14 failures on the two-solution tasks. Successes over trials with objects placed by hand. Time per query is measured during these trials on the robot's workstation (one RTX~4090).}
\label{tab:real}
\end{table}

The analysis of Sec.~\ref{sec:diagnosis} predicts that plans help most where a state admits several approaches, so on hardware the gain should appear on tasks with more than one solution rather than on a task with one.
We check this on a Franka arm (two cameras, 15\,Hz), fine-tuning the same backbone on teleoperated demonstrations with two heads: \method at $K{=}8$, and OpenVLA-OFT+~\cite{oft} (L1 regression with FiLM), which outputs a single chunk (Table~\ref{tab:real}).
The one-step push has one solution, and the two policies are one trial apart (24/25 and 25/25).
The other two tasks each admit two solutions, along the two commitments of Sec.~\ref{sec:intro}: sub-goal order, since the blocks can enter the tray in either order, and approach direction, since the can may pass the obstacle on either side, though only 8 of the 58 demonstrations take the far side.
Here \method succeeds in 51 of 54 trials against 40, removing 11 of the baseline's 14 failures.
On the blocks the baseline twice hovers without committing to its next motion, and on the can task it displaces the obstacle in 8 trials and is clean in 1 of its 6 far-side passes; \method takes both orders and both sides, and all 5 of its far-side passes are clean.
Eight candidates do not slow the robot down: 113\,ms per query against 145\,ms for OFT+.
To isolate the plans on the robot model as well, we repeat the same-checkpoint contrast of Table~\ref{tab:mechanism}a offline, scoring candidates against held-out demonstrations instead of executing them: eight plan-conditioned candidates recover 19.6\% of the first candidate's misses (errors above half the demonstrated displacement), against 7.1\% for eight same-z samples.
The plan statistics agree: on the two-solution tasks an episode on average draws on 7 of the 8 plans and its most used plan serves only 37--47\% of the decisions, whereas on the push a single plan serves 74\%.

\section{Conclusion}
\label{sec:conclusion}

Test-time scaling for VLAs has mostly improved how candidates are scored; we measured under one recipe how much the candidate source limits it, and raised that limit by making the candidates differ by plan.
Resampling one decoder has quickly diminishing returns: three resampling sources gain only 0.06--0.07 completed tasks from four to eight candidates, and even an oracle selector finds a successful candidate among eight same-plan resamples in only 86\% of sub-task start states.
\method decodes its candidates from $K$ latent plan hypotheses, learned from demonstrations alone through latent responsibility assignment, and serves any candidate budget up to $K_{\max}$ from a single checkpoint.
At matched candidate counts \method outperforms autoregressive, plan-free and same-plan resampling on CALVIN ABC$\rightarrow$D and LIBERO-Long, removing 30\% and 40\% of single-candidate failures against at most 21\%, without any external model and at a fraction of the latency of autoregressive sampling.
In a fixed-noise intervention, changing only the plan alters the task outcome more than twice as often as changing only the decoder noise, and on a real Franka arm \method removes 11 of the 14 failures of an OpenVLA-OFT+ baseline on tasks with more than one solution.

\noindent\textbf{Limitations and outlook.}
Our study uses one backbone and one arm; richer executors~\cite{ddvla} and stronger selectors can be placed on top of plan-conditioned candidates.

\section*{LLM Usage Statement}
The research ideas, the method, the experiments and the claims of this paper are the authors' own.
Claude (Anthropic) assisted the authors, under their direction, in three ways: (i) literature search, finding and summarizing related work; (ii) code, debugging and fixing code and writing scripts that check the reported numbers; and (iii) the manuscript, drafting text from the authors' notes, polishing and revising the text of all sections, preparing the \LaTeX{} source in the conference template (including the code for tables and figures), and giving review-style feedback on drafts.
The authors take full responsibility for the content.

{
    \small
    \bibliographystyle{ieeenat_fullname}
    \bibliography{main}
}

\end{document}